%% file: root.tex
\documentclass[letterpaper, 10 pt, conference]{ieeeconf}  % Comment this line out if you need a4paper

\IEEEoverridecommandlockouts                              % This command is only needed if 
\usepackage{amsmath, amsfonts}
\usepackage{xcolor}
\usepackage{float}
\usepackage{graphicx}
\usepackage{wrapfig}
\usepackage{cite}
\usepackage{svg}
\let\labelindent\relax
\usepackage{enumitem}
\usepackage{url}
\usepackage{hyperref}

\newcommand{\bx}{\pmb{x}}

\newcommand{\bw}{\pmb{w}}

\newcommand{\mP}{\mathcal{P}}
\newcommand{\rar}{\rightarrow}

\title{\LARGE {\bf
DiffCloud:} Real-to-Sim from Point Clouds with \\    Differentiable Simulation and Rendering of Deformable Objects
}
\author{Priya Sundaresan$^{1\dagger}$, Rika Antonova$^{1*}$, and Jeannette Bohg$^1$% <-this % stops a space
\thanks{$^{1}$Department of Computer Science, Stanford University, Stanford, CA 94305, USA
        {\tt\scriptsize \{priyasun, \ rika.antonova, \ bohg\}@stanford.edu}}%
\thanks{$^\dagger$P. Sundaresan was supported by the NSF Graduate Research Fellowship.}
\thanks{$^*$Supported by the National Science Foundation grant No.2030859 to the Computing Research Association for the CIFellows Project.}
\thanks{This project was supported in part by a research award from Meta.
 The authors also thank Krishna Murthy Jatavallabhula for helpful discussions.}
}

\definecolor{mydarkblue}{rgb}{0,0.08,0.45}
\hypersetup{ %
    pdftitle={DiffCloud: Real-to-Sim from Point Clouds with Differentiable Simulation and Rendering of Deformable Objects},
    pdfauthor={Priya Sundaresan, Rika Antonova, Jeannette Bohg},
    pdfsubject={Robotics},
    pdfkeywords={Differentiable Simulation, Robot Manipulation},
    pdfborder=0 0 0,
    pdfpagemode=UseNone,
    colorlinks=true,
    linkcolor=mydarkblue,
    citecolor=mydarkblue,
    filecolor=mydarkblue,
    urlcolor=mydarkblue,
    pdfview=FitH
}

\begin{document}

\maketitle
\thispagestyle{empty}
\pagestyle{empty}

%%%%%%%%%%%%%%%%%%%%%%%%%%%%%%%%%%%%%%%%%%%%%%%%%%%%%%%%%%%%%%%%%%%%%%%%%%%%%%%%
\begin{abstract}

Research in manipulation of deformable objects is typically conducted on a limited range of scenarios, because handling each scenario on hardware takes significant effort. Realistic simulators with support for various types of deformations and interactions have the potential to speed up experimentation with novel tasks and algorithms. However, for highly deformable objects it is challenging to align the output of a simulator with the behavior of real objects. Manual tuning is not intuitive, hence automated methods are needed.
We view this alignment problem as a joint perception-inference challenge and demonstrate how to use recent neural network architectures to successfully perform simulation parameter inference from real point clouds. We analyze the performance of various architectures, comparing their data and training requirements. Furthermore, we propose to leverage differentiable point cloud sampling and differentiable simulation to significantly reduce the time to achieve the alignment. We employ an efficient way to propagate gradients from point clouds to simulated meshes and further through to the physical simulation parameters, such as mass and stiffness. Experiments with highly deformable objects show that our method can achieve comparable or better alignment with real object behavior, while reducing the time needed to achieve this by more than an order of magnitude. Videos and supplementary material are available at \href{https://diffcloud.github.io}{{diffcloud.github.io}}.

\end{abstract}
%%%%%%%%%%%%%%%%%%%%%%%%%%%%%%%%%%%%%%%%%%%%%%%%%%%%%%%%%%%%%%%%%%%%%%%%%%%%%%%%

\input{1_intro}

\input{2_background}

\input{3_approach}
\input{4_experiments}

\input{5_conclusion}

\bibliographystyle{IEEEtran}
\bibliography{references}

\end{document}

%% file: 1_intro.tex
\section{Introduction}

We consider the \textit{real-to-sim} problem of inferring parameters of general-purpose simulators from real observations such that the gap between reality and simulation is reduced~\cite{prakash2021self, chang2020sim2real2sim, zhang2019vr, liu2021real}. A common approach to solving this problem is to train an inverse model on data generated with a black-box (non-differentiable) simulator using a wide range of parameter settings. The input to such models usually consists of trajectories of a low-dimensional state of the system, e.g. position and orientation of rigid objects in the scene. The output are parameters, such as mass, friction, and other physical object properties.

The state of a highly deformable object cannot be captured by only its position and orientation, since the object deforms during motion. Hence, we represent objects with point clouds as observed through depth cameras. 
Recent neural network architectures, such as PointNet++~\cite{qi2017pointnet++} and MeteorNet~\cite{liu2019meteornet}, are well suited for learning to process point clouds. As we will show, they can yield inverse models that offer a viable solution to the challenging task of real-to-sim for deformables from point clouds. Nonetheless, data collection and training for these can be computationally demanding.
%This design choice typically leads users to separate the problem into two stages: using external computer vision techniques to infer the poses of objects in the scene, then using differentiability in the simulator to optimize simulator or control parameters.

%The recent success of differentiable approaches in machine learning brought a heightened interest in differentiability to robotics. 

\begin{figure}[t]
    \centering
    \includegraphics[width=0.48\textwidth]{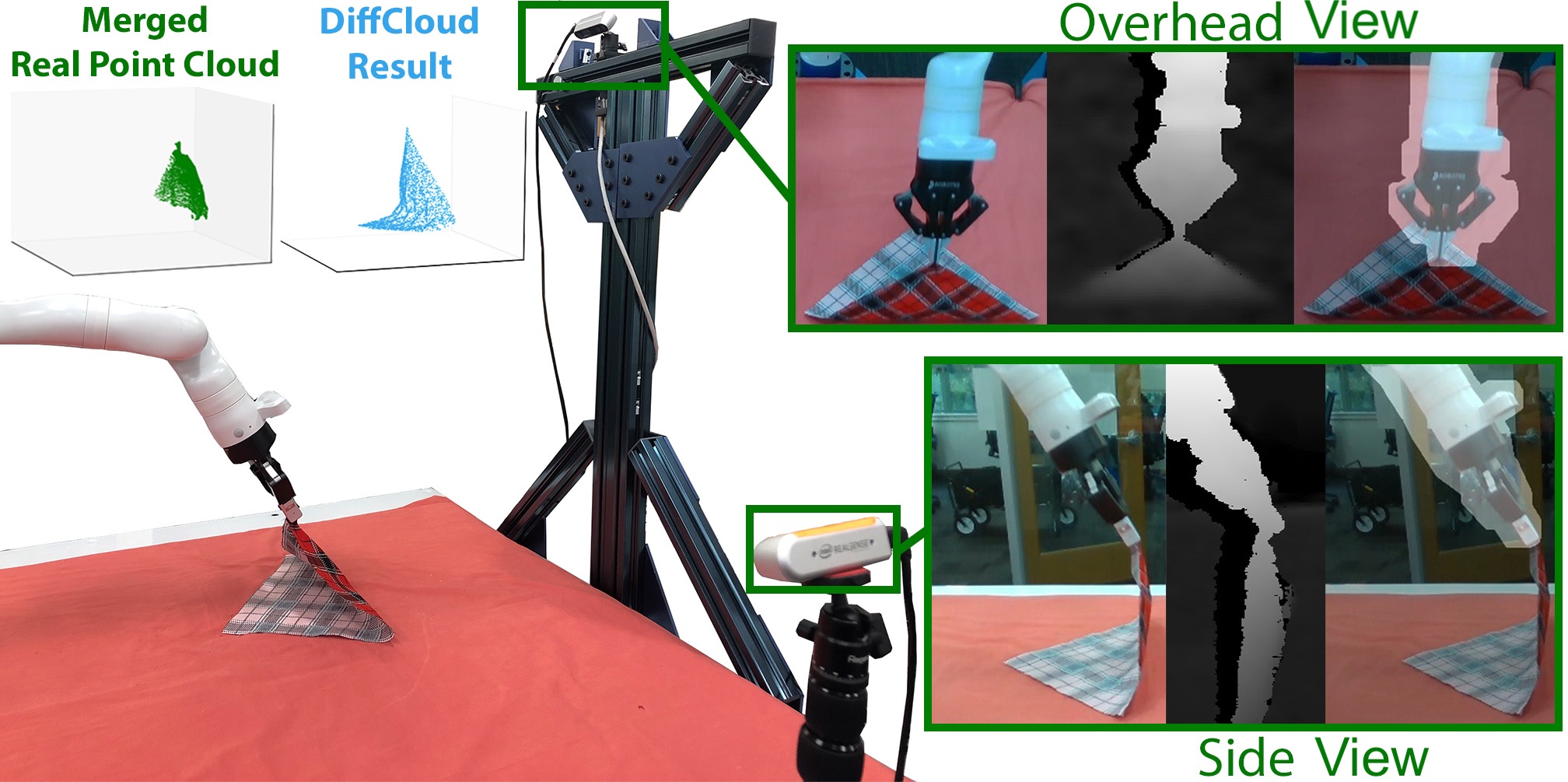}
    \vspace{-20px}
    \caption{Experimental setup. We execute deformable manipulation trajectories using a Kinova Gen3 arm. We post-process observations recorded from two stereo depth cameras (Intel D435) to generate merged point clouds with the robot arm masked from view. These observations are fed to our proposed method \textsc{DiffCloud} for \textit{real-to-sim} parameter estimation.}
    \label{fig:splash_fig}
    \vspace{-15px}
\end{figure}

In this work, we propose an alternative approach that employs a differentiable simulator to allow adjusting simulation parameters directly via gradient descent without the need for dataset collection and pre-training. Our approach combines differentiable point cloud rendering and differentiable simulation to bring the behavior of simulated, highly deformable objects closer to that of real objects. 
We instantiate a scene with a simulated object and create an end-to-end differentiable pipeline that lets us seamlessly propagate the gradients from real point clouds to the low-level physical simulation parameters. For highly deformable objects, even small changes in these parameters can have a significant impact on the behavior of the object. We show that establishing end-to-end differentiability yields a faster alignment between simulation and reality, compared to training inverse models with a black-box (i.e. gradient-free) treatment of the simulator.

We conduct a set of experiments where a robot manipulates highly deformable real objects, such as cloth and paper towels. We show that our approach successfully infers simulation parameters, such as mass and stiffness, making the behavior of simulated deformables match the real ones. In simulation experiments, we explore interactions of the deformables with rigid objects: stretching a band on a pole, and hanging a vest onto a rigid pole. Overall, our experiments show that we can obtain similar or better alignment between the simulated and the target object, compared to the inverse model baselines. The major benefit of our approach is that it obtains such an alignment after on average \textit{10 minutes} of direct gradient descent, replacing \mbox{\textit{2.5 -- 5.5 hours}} of data collection and training for the inverse models.

%% file: 2_background.tex
\section{Background}
%\begin{figure*}[t]
%\centering
%\includegraphics[width=1.0\textwidth]{imgs/fig1.png}
%\caption{\textbf{Overview of \textsc{DiffCloud}}: the proposed method for real-to-sim parameter estimation from point clouds. \textsc{DiffCloud} uses a differentiable point cloud sampling procedure combined with an end-to-end differentiable mesh-based simulator to propagate losses between observed simulated and real point clouds to the underlying properties of simulated cloth. We visualize the pipeline for updating mass and stiffness of a simulated 3D vest hanged on a pole.}
%\label{fig:method_overview}
%\end{figure*}
% TODO(Priya): address Jeannette's comments about the overview figure. (edit: done)
%---------------------------------------
\subsection{Real-to-Sim for Deformable Objects}
Many approaches in deformable object manipulation are limited to specific scenarios due to the complexity of real hardware setups~\cite{yoshida2015simulation,matas2018sim,klee2015personalized,kapusta2019personalized,clegg2018learning,shen2021provably}. Furthermore, there is a lack of easily tunable yet realistic simulators that support deformables and could aid experimentation with novel tasks.
%consider only a restricted range of scenarios.
%Some recent examples include (in separate works): manipulating an elastic loop~\cite{yoshida2015simulation}; hanging a piece of cloth~\cite{matas2018sim}; assisting to put on a hat~\cite{klee2015personalized}, a shirt~\cite{clegg2018learning}, a gown~\cite{kapusta2019personalized}, a sleeveless jacket~\cite{shen2021provably}. Setting up these scenarios on hardware takes significant effort. Hence, realistic simulators with support for various types of deformable objects is a much needed aid to speed up experimentation with novel tasks as well as perception and manipulation algorithms. A major obstacle is that such simulators are difficult to tune manually, because of the interplay between various physical parameters that impact the deformation, and because it is difficult to see at a glance whether a highly deformable, simulated object undergoes the same deformations as the real one. 
Hence, automated ways to find simulation parameters are needed to make the behavior of simulated deformables resemble that of their real-world counterpart. In robotics, this \textit{real-to-sim} problem has been extensively studied for rigid objects~\cite{chebotar2019closing, ramos2019bayessim, barcelos2020disco, mehta2020active, muratore2021neural, hwasser2020variational}. However, these methods assume access to a low-dimensional state, such as object poses. Recent surveys have reviewed learning-based approaches for manipulation and perception of deformables~\cite{herguedas2019survey, arriola2020modeling, yin2021modeling}, including methods for tracking and registration. However, such methods either employ markers~\cite{navarro2013model}, assume a known and simplified deformation model~\cite{sun2008learning}, or show applicability only to specific cases, e.g. a rope lying flat on a surface~\cite{tang2018framework}. As of now, there is no generic tracking or registration approach for deformables that is robust in a wide range of scenarios. %Hence, for deformables there a is lack of reliable and general perception methods for mapping from camera images or point clouds to a mesh or low-dimensional state. %As a consequence, it is difficult to split the problem into separate perception and parameter inference stages.
%Dealing with deformable objects requires developing methods that can natively handle high-dimensional inputs, such as point clouds and RGB images. This is because it is not easy to manually define a low-dimensional representation for the state of deformables, especially for the highly deformable objects, such as cloth and ropes. 

%---------------------------------------
%\subsection{Inverse Models for Combined Perception and Inference}
\subsection{Inverse Models for Real-to-Sim}
One approach towards real-to-sim is to learn an inverse model that 
maps from a sequence of sensor observations to simulation parameters~\cite{willard2020integrating}. In robotics, several earlier works explored using inverse models for highly deformable objects. For example, in~\cite{wang2011data} the authors propose a piecewise linear elastic material model for cloth, then they fit the material model to real cloths by applying controlled forces and measuring the deformation response. This results in a paired dataset of cloth types and estimated material parameters, but limits generalization to unseen cloths without additional physical measurements. In \cite{yang2017learning}, the authors propose using the dataset of \cite{wang2011data} as a source of supervision for training a network to regress material parameters from RGB videos of cloth. This supervised training approach relies heavily on domain randomization to achieve generalization to cloths with unseen colors or patterns. Additionally, the proposed regression framework discretizes the space of outputs to deal with the high dimensional model of cloth deformation, limiting the range of cloth types that can ultimately be accurately modelled.
In this work, we use a sequence of depth images (converted to point clouds) as input to our models. Using point clouds instead or RGB images allows us to circumvent the need for extensive domain randomization and for contrasting textures (for foreground/background segmentation). 
We adopt established methods for processing point clouds, such as PointNet/PointNet++~\cite{qi2017pointnet,qi2017pointnet++}, MeteorNet~\cite{liu2019meteornet} and train them to regress to continuous simulation parameters.
\subsection{Differentiable Rendering and Differentiable Simulation}

{\em Differentiable rendering\/} of images allows using 2D observations to inform 3D understanding, by propagating gradients from images to illumination models, 3D objects in the scene, and camera pose.
%~\cite{loper2014opendr,insafutdinov2018unsupervised,chen2019learning,liu2019soft,lassner2021pulsar,El_Banani_2021_CVPR,planche2021physics}. 
Various works have explored making the rasterization process (coloring image pixels based on the assignment of triangular mesh faces) fully differentiable, including DIB-R~\cite{chen2019learning} and SoftRAS~\cite{liu2019soft}. Other works explore making the processes of ray tracing~\cite{tulsiani2017multi} and volumetric rendering~\cite{niemeyer2020differentiable, mildenhall2020nerf} differentiable. Toolkits, such as Kaolin~\cite{jatavallabhula2019kaolin} and PyTorch3D~\cite{ravi2020pytorch3d} provide batched implementations of differentiable rasterization, allowing a significant speedup of these computationally expensive processes. In this work, we aim to do correspondence-free alignment of simulated object states, such as meshes, with real visual observations. 
While images are one such observation space to bridge the gap, achieving close alignment between images that are differentiably rendered from simulated meshes and real images still remains an open challenge due to the range of colors, textures, and fine details present in real images. Thus, we use point clouds to represent objects and build on PyTorch3D~\cite{ravi2020pytorch3d}, which offers a way to differentiably sample a point cloud from a given mesh such that gradients from losses in the point clouds propagate to the mesh vertex gradients. Furthermore, PyTorch3D library exposes efficient data structures and CUDA-enabled batched operations for 3D meshes and point clouds, allowing seamless integration into simulator backends that support PyTorch.

\begin{figure*}[t]
\centering
\vspace{10px}
\includegraphics[width=1.0\textwidth]{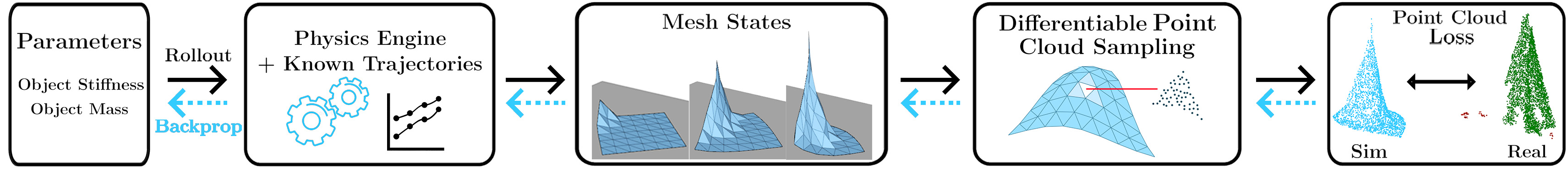}
\vspace{-22px}
\caption{\textbf{Overview of \textsc{DiffCloud}}: the proposed method for real-to-sim parameter estimation from point clouds. \textsc{DiffCloud} combines differentiable point cloud sampling with a differentiable mesh-based simulator to propagate losses computed between simulated (blue) and real (green, with red noise artifacts) point clouds to the underlying simulation properties. We visualize updating mass and stiffness of a simulated cloth lifted off of a table.}
\vspace{-12px}
\label{fig:method_overview}
\end{figure*}

\textit{Differentiable simulation} is a promising paradigm that allows to adapt simulators by propagating gradients of the simulator outputs w.r.t. the lower-level physical simulation parameters. 
%This kind of optimization can be more direct and yield more precise real-to-sim alignment than treating a simulator as a black box. 
Recent differentiable simulators with support for deformables include~\cite{qiao2020scalable,krishna2021gradsim,hu2019difftaichi,liang2019differentiable,hu2019chainqueen,hahn2019real2sim,geilinger2020add,du2021diffpd,lutter2021differentiable,mora2021pods}.
Most of these only support a limited set of interactions and types of deformation. For example, out of the above, only~\cite{qiao2020scalable} (a differentiable version of ARCSim~\cite{narain2012adaptive}) supports arbitrary meshes and modeling interactions of rigid and deformable objects.
From the simulators listed above, only gradSim~\cite{krishna2021gradsim} provides a combination of differentiable RGB rendering and simulation. This approach is conceptually related to our work, but is still substantially different, since it does not handle point clouds, does not support interactions between rigid and deformable objects, and presents only sim-to-sim results with simple simulated images and plain textures.
The differentiable simulation framework proposed in \cite{weiss2020correspondence} shows results on inferring deformation properties of a real object from point clouds. We do not build on this approach directly, because it has several key limitations. First, there are limitations in the underlying simulator: the model is aimed for objects with low-to-medium levels of deformability, such as tight plush toys and pillows. In contrast, we are interested in highly deformable objects, such as cloth, garments, and stretchable bands. Second, there are limitations in the loss formulation: the point cloud loss proposed in~\cite{weiss2020correspondence} relies on a simple diffusion of the observed deformation into a deformation field. This is not suitable for the case of highly deformable objects that rapidly and drastically change shape under gravity and upon interacting with other objects. Finally, this prior work is geared towards analyzing cases with passive dynamics of a single object falling and colliding with the table or bending under gravity; the simulator also lacks support for modeling frictional contacts. We are instead interested in actuating a robot to perform manipulation with highly deformable objects that can also interact in non-trivial ways with rigid objects in the scene. We found that~\cite{qiao2020scalable} is the only available simulator that readily supports such functionality, hence we incorporate it into our approach.

%% file: 3_approach.tex
%\begin{figure*}[t]
%\centering
%\vspace{10px}
%\includegraphics[width=1.0\textwidth]{imgs/system_overview.jpg}
%\vspace{-22px}
%\caption{\textbf{Overview of \textsc{DiffCloud}}: the proposed method for real-to-sim parameter estimation from point clouds. \textsc{DiffCloud} combines differentiable point cloud sampling with a differentiable mesh-based simulator to propagate losses computed between simulated (blue) and real (green, with red noise artifacts) point clouds to the underlying simulation properties. We visualize updating mass and stiffness of a simulated cloth lifted off of a table.}
%\vspace{-12px}
%\label{fig:method_overview}
%\end{figure*}

\section{Our Approach : DiffCloud}
\label{sec:diffcloud}

Our objective is to discover the physical simulation parameters (e.g. stiffness, mass, friction) that would cause the behavior of the simulated deformable objects to match the behavior of observed real objects. We start by recording a sequence of point cloud observations of the scene, where a robot manipulates a deformable object. Then, we construct a simulated environment in a differentiable simulator that can load meshes of deformable \& rigid objects in the scene and simulate their interactions.
%For example, differentiable simulator from~\cite{qiao2020scalable} supports mesh representations and interactions between rigid and deformable objects. 
To obtain simulated point clouds we implement a differentiable point cloud sampler: we sample points on the surface of the simulated objects using PyTorch3D~\cite{ravi2020pytorch3d}. By defining a differentiable loss between the simulated and real point cloud sequences we can backpropagate all the way to the simulation parameters. Figure \ref{fig:method_overview} gives an overview of our approach.

In this work, we assume that the initial geometry of objects in the scene is known. This allows us to initialize the simulation start state to match the real scene. We also assume that the location of the grasp is known, so we can grasp the object in the simulated scene in the same way as in reality (with a simulated gripper or gripper anchors). With that, we can control the simulated gripper (or anchors) along the trajectory of the end effector that was recorded in reality.

%---------------------------------------
\subsection{Loss Definition}
\label{sec:loss}

One candidate for a loss function on point clouds is the Chamfer distance:
\begin{align}
\label{eq:chamfer}
\begin{split}
d_{\text{Chamf}}(\mP^a, \mP^b) \!=& \tfrac{1}{|\mP^a|} \! \sum_{\bx^a \in \mP^a}\!\! \min_{\bx^b \in \mP^b} || \bx^a - \bx^b ||_2^2 \\
&+\! \tfrac{1}{|\mP^b|} \! \sum_{\bx^b \in \mP^b}\!\! \min_{\bx^a \in \mP^a} || 
\bx^b - \bx^a ||_2^2.
\end{split}
\end{align}
Here, $\mP^a, \mP^b$ denote two point clouds; $\bx^a, \bx^b$ are 3D points in $\mP^a$ and $\mP^b$ respectively. This distance metric is frequently used to compare the alignment between two point clouds that are either complete (sampled from meshes) or are generated by perception with the same camera perspective and noise properties.
In our case of using low-cost depth sensors, the challenge of constructing the loss function on the real and simulated point clouds is that we need to avoid paying attention to the noise artifacts in the real point cloud. Statistical outlier filtering and post-processing techniques can alleviate noise, but some amount is likely to remain. For example, see the red patches highlighted in the real point cloud in Figure~\ref{fig:splash_fig}.
Our insight is that a loss that disregards noise artifacts can be obtained by using a unidirectional Chamfer distance. This yields a loss that relieves the pressure for the simulated point clouds to match the noisy parts of the real point clouds:
\begin{align}
\vspace{-5px}
\label{eq:unidir_chamfer}
\begin{split}
d_{\text{Chamf}}^{\text{sim} \rar \text{real}}(\mP^{\text{sim}}, \mP^{\text{real}}) \!=\!\! \sum_{\bx^{\text{sim}} \in \mP^{\text{sim}}}\!\! \min_{\bx^{\text{real}} \in \mP^{\text{real}}} || \bx^{\text{sim}} - \bx^{\text{real}} ||_2^2.
\end{split}
\end{align}

While the naive computation of the Chamfer distance can be expensive, PyTorch3D provides an efficient GPU-based implementation (see Section 3.1 in~\cite{ravi2020pytorch3d}).
With that, we can quickly propagate gradients from the point clouds through the mesh representation to optimize the low-level physical parameters of the simulator. We found that including point clouds from one or two depth cameras is sufficient to construct a partially occluded point cloud that is still informative enough for the overall optimization to be successful.

\subsection{Gradient Propagation}
\label{sec:grad_prop}

We build upon the differentiable simulator DiffSim~\cite{qiao2020scalable}, which supports mesh-based simulation and contact-handling of rigid objects and thin-shell deformables (e.g. cloth). In DiffSim, the simulation state is represented by generalized coordinates $\mathbf{q} = [\mathbf{q_1}^T, \mathbf{q_2}^T, \hdots, \mathbf{q_n}^T]^T$ of all objects in the simulation with corresponding velocities $\mathbf{\dot{q}} = [\mathbf{\dot{q}_1}^T, \mathbf{\dot{q}_2}^T, \hdots, \mathbf{\dot{q}_n}^T]^T$. The generalized coordinates of a rigid body have $\mathbf{q_i} \in \mathbb{R}^6$ for rigid objects, denoting position and orientation, and $\mathbf{q_i} \in \mathbb{R}^3$ for deformable nodes with 3 DoF for position alone. Here, $n$ is the cumulative total of the number of deformable nodes and rigid bodies in the scene. DiffSim uses the implicit Euler method to compute $\mathbf{q, \dot{q}}$ at each time step and performs collision resolution in localized impact zones. A given mesh has a body frame with the origin set to its center of mass at the start of simulation. A mesh vertex $p$ has coordinate $\mathbf{p_0}  = (p_x, p_y, p_z)^T$ in the body frame and world coordinate $\mathbf{p} = \mathbf{f}(\mathbf{q})= [\mathbf{r}]\mathbf{p_0} + \mathbf{t}$, where $\mathbf{r}=(\phi, \theta, \psi)^T$ and $\mathbf{t}=(t_x,t_y,t_z)$ is the 6-DoF pose of the mesh. Propagating gradients from vertex $\mathbf{p}$ to the generalized coordinates $\mathbf{q}$ involves computing the Jacobian $\nabla\mathbf{f}$ and obtaining the partial derivatives $\partial{\mathbf{f(q)}}/\partial{\mathbf{q}}$. In this way, DiffSim is able to solve sim-to-sim optimizations by comparing current mesh states to target mesh states and propagating gradients from observed positional differences.

Since ground truth mesh states are not readily available for real deformables, we use point clouds as an observation space. To allow losses in the point cloud space to be propagated to simulated cloth vertices, we implement a differentiable point cloud sampling step. A triangular mesh face can be represented by its enclosing vertices $\mathbf{p_1}, \mathbf{p_2}, \mathbf{p_3}$. In barycentric coordinates, a random point on the surface of the face can be generated by sampling 3 random numbers $(u,v,w)$ such that $u + v + w \!\leq\! 1$~\cite{ravi2020pytorch3d}. 
%In practice, we randomly select $u,v$ and let $w = 1-u-v$. 
Hence, we can obtain a random point $\bx$ inside the triangle as:
\setlength{\abovedisplayskip}{3pt}
\setlength{\belowdisplayskip}{1pt}
\setlength{\abovedisplayshortskip}{3pt}
\setlength{\belowdisplayshortskip}{1pt}
\begin{equation}
\label{eq:barycen}
    \bx = u\mathbf{p_1} + v\mathbf{p_2} + w\mathbf{p_3}.
\end{equation}

To generate an $N$-point, uniform-density point cloud from a mesh, we first sample $N$ triangular faces $\{(\mathbf{p_{i,1}, p_{i,2}, p_{i,3}})\}_{i=1,\hdots,N}$,  weighted proportionally to the area of each face. This step ensures that the resulting point cloud is evenly distributed over the surface of the mesh, instead of being disproportionately dense in regions where the mesh has closely packed faces. For the $i^{\text{th}}$ sampled face $(\mathbf{p_{i,1}, p_{i,2}, p_{i,3}})$, we generate random coefficients $(u_i, v_i, w_i)$ and use Equation \ref{eq:barycen} to compute a point $\bx_i$ that lies on the face. Concatenating the points obtained by applying this procedure to the $N$ sampled faces and coefficients yields the point cloud $\mathcal{P} \!=\! \{\bx_i\}_{i=1,\hdots,N}$. We connect the PyTorch3D~\cite{ravi2020pytorch3d} implementation of this sampling procedure to the output of a differentiable simulator. With that, gradients from loss functions operating on $\{\bx_i\}_{i=1,\hdots,N}$ can be propagated to mesh vertices $(\mathbf{p_{i,1}, p_{i,2}, p_{i,3}})$ via chain rule, observing that:
\setlength{\abovedisplayskip}{3pt}
\setlength{\belowdisplayskip}{1pt}
\setlength{\abovedisplayshortskip}{3pt}
\setlength{\belowdisplayshortskip}{1pt}
\begin{equation}
\label{eq:partials}
    \partial{\bx_i}/\partial{\mathbf{p_{i,1}}}=u_i \hspace{0.3cm} \partial{\bx_i}/\partial{\mathbf{p_{i,2}}}=v_i \hspace{0.3cm} \partial{\bx_i}/\partial{\mathbf{p_{i,3}}}=w_i
\end{equation}

\subsection{Optimization}
We consider scenarios where a robot executes a trajectory to manipulate a deformable object, which potentially also interacts with other objects in the scene. 
For each scenario, we get a real point cloud sequence of length $T$ and a resolution of $N$ points per frame: $\big\{\mathcal{P}^{\mathrm{real}}_t \!=\! \{\bx_i^{\text{real}}\}_{i=1,\hdots,N}\big\}_{t=1}^T$.
We instantiate each real scenario in DiffSim using representative meshes to model the deformable and rigid objects in the scene initially. Next, we begin optimizing the material properties of the simulated cloth such that the discovered parameters best explain the observed target point cloud data. Each simulated scenario involves manipulating a simulated cloth with the same trajectory as in real. Even with identical trajectory execution, we expect a discrepancy between the real and simulated point cloud sequences due to mismatch in the motion of the deformables, whose physical properties we aim to optimize. For each iteration of optimization, we run the simulation for the number of steps in the scenario horizon using the current set of estimated parameters. At each timestep $t$, we sample the simulated object surfaces in the scene according to Equation \ref{eq:barycen} to obtain a point cloud. The concatenated point clouds across frames give the simulated point cloud sequence:
$\big\{\mathcal{P}^{\mathrm{sim}}_t = \{\bx_i^{\text{sim}}\}_{i=1,\hdots,N}\big\}_{t=1}^T$.
In practice, we compute the unidirectional Chamfer distance (Equation \ref{eq:unidir_chamfer}) on $(\mathcal{P}^{\mathrm{sim}}_t, \mathcal{P}^{\mathrm{real}}_t)$ for the corresponding point clouds observed at time $t$. For most experiments, we found that using only the last frame works well, i.e. $t=T$, as in \cite{krishna2021gradsim, qiao2020scalable}. Other frames can be used, and their selection could be considered as a hyperparameter. This loss is propagated across all frames of the simulation and used to update the underlying parameters of the deformable object. This procedure is repeated for a fixed number of iterations or until the Chamfer distance falls below a threshold.

%% file: 4_experiments.tex
\section{Experiments}
\label{sec:experiments}
In this section, we compare \textsc{DiffCloud} with non-differentiable methods for parameter estimation: we evaluate the degree of point cloud alignment between simulated and real trajectories and the compute efficiency across methods.
%In this section we compare \textsc{DiffCloud} with methods that employ various point cloud architectures while treating simulation as a non-differentiable source of training data. We aim to answer the following questions:
%\begin{itemize}
%\item[1)] How well does \textsc{DiffCloud} perform compared to NN-based inverse models for finding simulation parameters that make the real and simulated object motion similar? 
%\item[2)] How much speedup does \textsc{DiffCloud} provide via direct gradient-based optimization, compared to data collection and supervised training of the inverse models?
%\end{itemize}

\begin{figure}[t]
    \vspace{10px}
    \centering
    % \includegraphics[width=0.47\textwidth]{imgs/times_params.png}
    % \includesvg[width=0.48\textwidth]{imgs/times_params.svg}
    \includegraphics[width=0.48\textwidth]{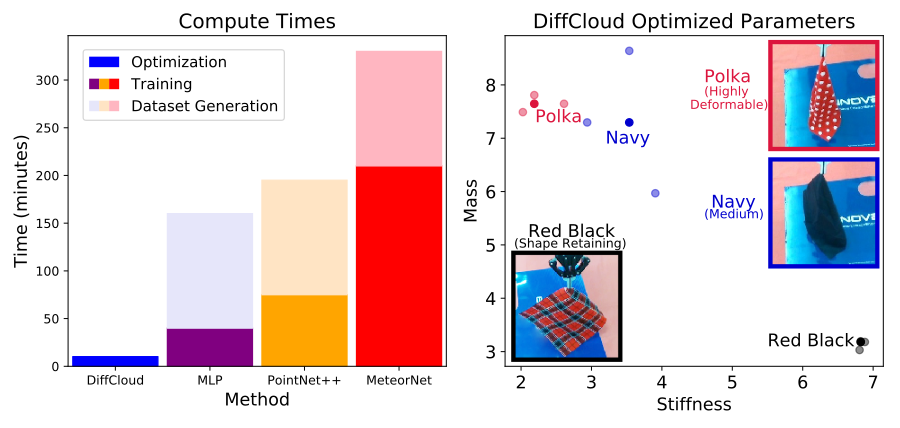}
    \vspace{-25px}
    \caption{Left: We visualize the average compute times across all methods for performing parameter estimation in the real lift and fold scenarios. Compared to the baseline inverse models, \textsc{DiffCloud} achieves more than an order of magnitude speedup, since it eliminates the need to pre-generate a dataset and train on it. Right: The optimized \textsc{DiffCloud} parameters found in the lift scenario correspond to the intuitive physical properties of real cloths, ranging from highly deformable to shape retaining. Each darkened circle represents the category median across three trajectories per cloth type.}
    \label{fig:method_timings_params}
    \vspace{-15px}
\end{figure}

% ----------------------------------
\subsection{Baseline Inverse Models}
As experimental baselines, we use methods that view simulators as black-box, i.e. not treating observations as end-to-end differentiable w.r.t the parameters.
These inverse models are implemented as regression networks that take point cloud sequences as inputs and predict $k$ simulation parameters. They are trained on simulated point cloud sequences generated by simulations with various simulation parameters. 

\begin{itemize}[leftmargin=*]
\item \textsc{MeteorNet}: An architecture for learning representations of 3D point cloud sequences from~\cite{liu2019meteornet}, which we modify to predict $k$ simulation parameters. Specifically, we use MeteorNet-cls (Appendix D.2 in~\cite{liu2019meteornet}).

% {\color{magenta}TODO[Priya]: add NN sizes and training}

\item \textsc{PointNet++}: A regressor similar to the above, but using the multi-scale group architecture from~\cite{qi2017pointnet++} (Appendix B.1) to extract features from a single point cloud; uses three set abstraction layers followed by three fully connected layers with output sizes $(512, 256, k)$.
% {\color{magenta}TODO[Priya]: add NN sizes}.

\item \textsc{MLP}: A regressor with a fully connected network that also operates on a single frame; uses five 2D convolutional layers with output sizes $(64,64,64,128,1024)$, a symmetric max pooling layer, two fully connected layers with output sizes $(512, 256)$, a dropout layer, and a final fully connected layer with $k$ outputs.

\end{itemize}

Real point cloud sequences suffer from partial observability, self-occlusion, and noise. To minimize the gap between simulation and reality, we aim to generate a training dataset for each of the regressor methods that is as realistic as possible. As training data for the regressors (\textsc{MeteorNet}, \textsc{PointNet++}, \textsc{MLP}), we initialize $N$ simulations with uniformly sampled parameters and record the resulting point cloud sequences and ground truth parameters. This yields a dataset $\mathcal{D} = \big\{ \{\bx_t\}_{t=1}^T, \bw \!=\![w_{\mathrm{stiff}}, w_{\mathrm{mass}}] \big\}_{i=1}^N$.
For each simulation run, we generate a point cloud sequence from mesh states according to Equation~\ref{eq:barycen}, but bias the sampling of points to be on a subset instead of all faces of the deformable to mimic a partial point cloud with occlusion. Then, we apply random Gaussian noise to this point cloud to mimic the effect of real sensor noise. We generate a dataset of 1500 training and 375 test point cloud sequences with 3,500 points per frame across methods. For comparison, \textsc{MeteorNet} was trained on a dataset of 576 examples and the PointNet++~\cite{qi2017pointnet++} literature uses thousands of examples.

We train the \textsc{MeteorNet} regressor to learn a mapping $f: \{\bx\}_{i=1}^T  \rar \bw$, which maps an input point cloud sequence $\{\bx_t\}_{t=1}^T$ to simulator parameters $\bw$ that would generate the observed behavior. \textsc{PointNet++} and \textsc{MLP} methods learn a function $g: \bx_{t} \rar \bw$ mapping only a single frame $t$ (selected from a sequence) to the material parameters. In practice, we choose the frame $t$ to be the same frame for which we compute the unidirectional Chamfer distance in \textsc{DiffCloud} optimization. We train each network using \emph{$L_1$} loss between the predicted and ground truth parameters, using the Adam optimizer with a learning rate of 0.001 for 100 epochs.
We run dataset generation on an Intel i5-9400F CPU. We use an NVIDIA GeForce GTX 1070 GPU for training and optimization.

\subsection{\textsc{DiffCloud} Implementation Details}

\textsc{DiffCloud} is our proposed approach described in Section~\ref{sec:diffcloud}. For all simulations we use DiffSim~\cite{qiao2020scalable} -- a differentiable version of the garment simulator ARCSim~\cite{narain2012adaptive}. As in~\cite{yang2017learning}, we initialize the deformable, in this case cloth, to a basis material and learn two scalar multipliers $(w_\mathrm{stiff}, w_\mathrm{mass})$ for the stiffness and mass tensors. These multipliers represent the simulation parameters we aim to learn. We initialize them to the midpoint of each parameter range. We empirically determine this
range as $[0.1, 10]$ simulation units for all parameters, so that forward simulation is numerically stable. We want to encourage small changes in the learnable parameters to make a visually compelling difference in deformation.  
Thus, instead of directly optimizing $(w_\mathrm{stiff}, w_\mathrm{mass})$, we optimize intermediate variables $(s_\mathrm{stiff}, s_\mathrm{mass})$, where $(w_\mathrm{stiff}, w_\mathrm{mass}) = \sigma(s_\mathrm{stiff}, s_\mathrm{mass}) \times (10 - 0.1) + 0.1$, and $\sigma(x) = \frac{1}{1+e^{-x}}$ is the standard sigmoid function. Intuitively, this maps $(s_\mathrm{stiff}, s_\mathrm{mass})$ first to the range $[0,1]$ via the sigmoid function, and then interpolates this value within the parameter range of $[0.1,10]$. We choose the final parameters to be those that incurred the lowest loss during optimization.

\subsection{Evaluation Metrics}
\label{sec:evaluation}

\textit{Alignment: } For quantitative evaluation we use unidirectional Chamfer distance (Equation~\ref{eq:unidir_chamfer}) that characterizes how well the motions of the simulated and real point clouds align. For each of the baselines, we first infer the predicted parameters using the target point cloud trajectory as input. These parameters serve as input to the simulation engine. We then run the simulator using the inferred parameters and generate point clouds from the simulated meshes. We compute the Chamfer distance between the point clouds generated by the baselines and the real point cloud, then compare this against the loss from running \textsc{DiffCloud} on the real point cloud.

\textit{Efficiency: } Aside from alignment, we also evaluate the computational resources of all methods. Across experiments, we report (1) the time it takes to run \textsc{DiffCloud} optimization on trajectories and (2) the combined dataset generation, training, and inference times for the supervised baselines.

% For the \textsc{MLP, PointNet++}, and \textsc{MeteorNet}, we cannot optimize the Chamfer distance as a loss directly. This is because the only way to propagate gradients from the high-dimensional point cloud observation space to low-dimensional parameters would be via a decoding step. Decoding low-dimensional inputs into high-dimensional observation spaces can be problematic with non-trivial motion~\cite{antonova2020benchmarking}. The other option is to make the observation space end-to-end differentiable w.r.t the parameter space, which is precisely the motivation for \textsc{DiffCloud}.
% % This direct regression is more effective than decoder-based optimization, since decoding into high-dimensional spaces can be problematic with non-trivial motion.
% Hence, the best option for the baselines is to train networks that regress on the simulation parameters directly. This minimizes the discrepancies between real and simulated point clouds indirectly, but gives the baselines the best chance by presenting a tractable learning problem without the need for decoders. 

% ----------------------------------
\subsection{Real Experimental Setup}
\label{sec:real_exp_setup}

\subsubsection{Hardware Setup}

Our hardware setup consists of a Kinova Gen3 robot arm with a Robotiq 2F-85 gripper and two Intel RealSense D435 cameras (Figure \ref{fig:splash_fig}). The table workspace measures 50×43 cm, with one camera mounted overhead and another mounted on the side. For each scenario outlined below, we execute trajectories using velocity control in the Cartesian (end-effector) space and record robot joint states and point cloud observations at each step. We merge point clouds from both camera views and use a transformation obtained from standard checkerboard calibration to obtain point clouds in the frame of reference of the robot, which has a corresponding frame of reference in simulation. Using recorded joint states and known robot geometry, we mask out the end-effector from the point clouds as depicted in Figure \ref{fig:splash_fig}. This yields a set of real, post-processed point cloud observations $\{\bx{_t}^{\text{real}}\}_{t=1}^T$.

\subsubsection{Real Scenarios}

We consider two real manipulation scenarios using cloth as the deformable object of interest. The \texttt{lift} scenario involves grasping and lifting a cloth from a flat start state on a table surface. At the end of the trajectory, the cloth is lifted off the table and suspended mid-air. The \texttt{fold} scenario involves folding a cloth in half, starting from a diamond shape and ending in a triangular configuration. We perform both scenarios on 5 different fabrics grouped into three categories: highly deformable, medium, and shape retaining. The \textit{highly deformable} class consists of cloths that are especially low stiffness and collapsible, such as silk-like materials. The \textit{medium} class consist of cloths with modest deformation that can still crumple subject to enough force. The \textit{shape retaining} class contains cloths that resist deformation more than the other categories, such as paper towels and stiff washcloths. For each fabric, we execute 3 trajectories for a total of 15 trajectories per scenario. Each trajectory is $\approx\!2.5$ seconds long with robot commands sent at $10$ Hz., resulting in a horizon length of $T=25$. During the lift and fold trajectories, the cloth either collapses or maintains its shape depending on the underlying cloth properties (which we attempt to learn).

\begin{figure}[t]
    \vspace{10px}
    \centering
    \includegraphics[width=0.48\textwidth]{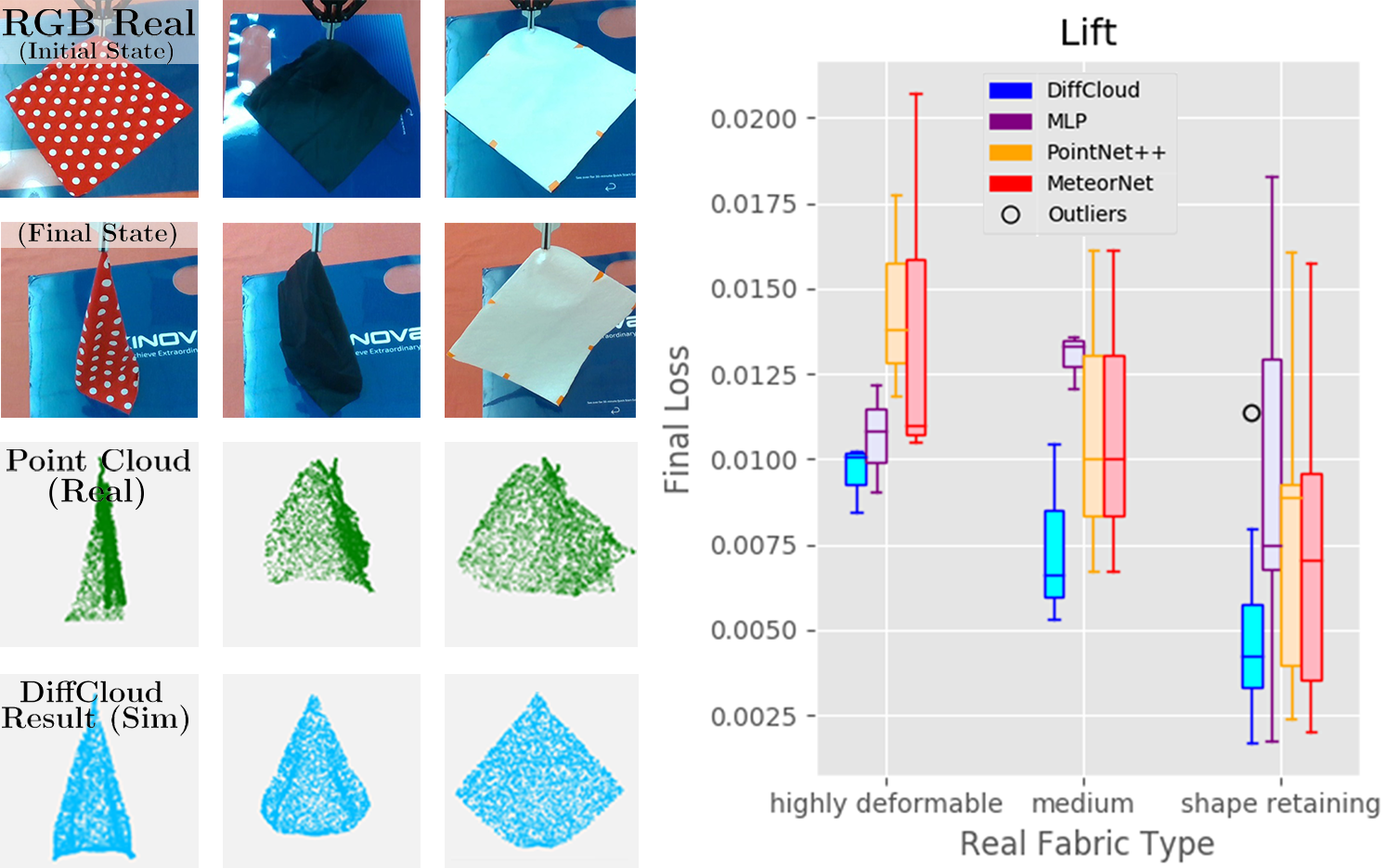}
    \vspace{-20px}
    \caption{Left: A Kinova robot executes trajectories to lift real cloths from a flat starting state. From these trajectories, \textsc{DiffCloud} accurately infers stiffness and mass parameters capturing the observed degree of collapsibility in the real cloths. Right: Across all cloth types, \textsc{DiffCloud} achieves lower loss on average than all competing baselines on the lift scenario. }
    \label{fig:lift}
    \vspace{-15px}
\end{figure}
% ----------------------------------
\subsection{Real Robot Experiments}
\label{subsec:real_exps}
% \begin{figure}
%     \centering
%     \includegraphics[width=0.5\textwidth]{imgs/params.png}
%     \caption{\textbf{Parameters Found by \textsc{DiffCloud}}: We visualize a box plot of the final stiffness and mass parameters found by \textsc{DiffCloud} across three fabric types. The parameters found for the trajectories for the paper category are low stiffness and low mass, which intuitively describe common kitchen paper towels. Similarly, the draping cloths exhibit a high learned mass and low learned stiffness which give the effect of a highly collapsible cloth as expected. The results for the medium deformation category, towels, is more varied as different combinations of stiffness and mass can result in moderate deformation.}
%     \label{fig:real_learned_params}
% \end{figure}

We aim to infer cloth stiffness and mass from the \texttt{lift} and \texttt{fold} scenarios discussed in Section \ref{sec:real_exp_setup}. 
  
\subsubsection{\textsc{DiffCloud Specifications}} 

% The real point cloud observations during both scenarios exhibit sensor noise and are especially lossy when the cloth is almost indistinguishable from the flat table initially. Thus, f
For specifying the \textsc{DiffCloud} loss and evaluation metrics, we need to use the frames for which deformation is most visible. The lift trajectory exhibits least occlusion at the end of the trajectory when the cloth is off the table. For folding, the cloth is most visible in the middle of the trajectory.
%due to the cloth laying flat at the beginning and ends of the fold. 
%Equation \ref{eq:unidir_chamfer} over the last and middle frames $S^{\text{c}} = \{\bx_{T}^{\text{i}}\}$ for lift and $S^{\text{c}} = \{\bx_{\lceil{\frac{T}{2}}\rceil}^{\text{c}}\}$ for fold, respectively, $c \in (\text{real, sim})$. 
The level of occlusion at the end of real lift trajectories is minimal, so simulated point clouds are generated by sampling from all cloth faces. In the fold scenario, the half of the real cloth that remains flat on the table does not appear in the real point cloud. Hence, we sample simulated point clouds only from cloth faces on the upper half of the cloth in simulation, as a heuristic for generating occlusion-sensitive point clouds. Occlusion handling is out of scope with the sampling procedure discussed in Section \ref{sec:grad_prop}, but we hope to relax this in future work. The simulated counterparts for the lift and fold scenarios both use a 2D cloth mesh consisting of a $7 \times 7$ grid of 49 vertices, where we keyframe the position of the top corner vertex to produce the same motion that is executed in real for $T=25$ steps. For both scenarios, we run \textsc{DiffCloud} for 50 iterations using the Adam optimizer with a learning rate of 0.2 to infer $(w_\mathrm{stiff}, w_\mathrm{mass})$.

\subsubsection{Results}

\begin{figure}[t]
    \vspace{10px}
    \centering
    \includegraphics[width=0.48\textwidth]{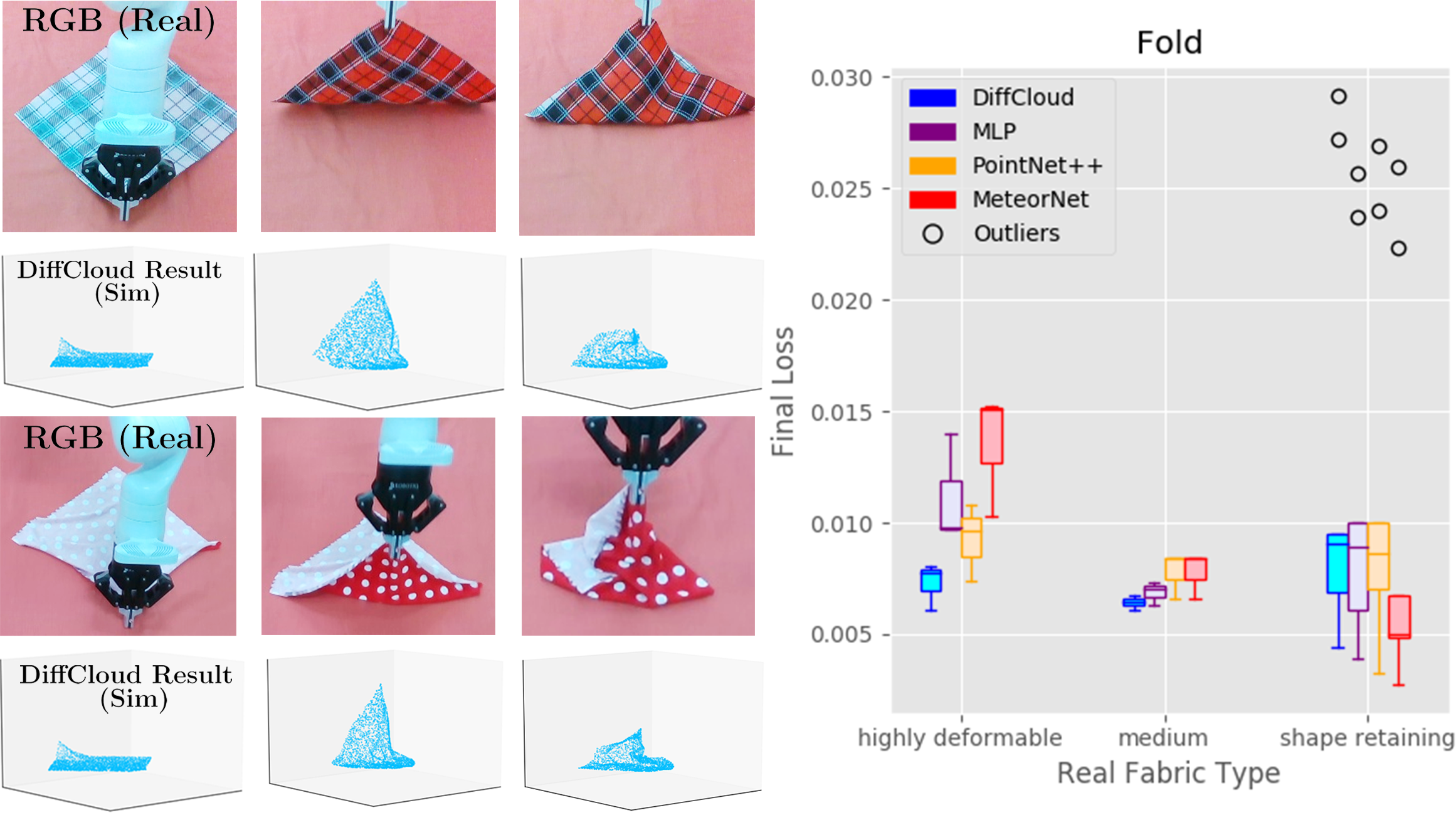}
    \vspace{-20px}
    \caption{Left: \textsc{DiffCloud} correctly learns to approximate low mass/high stiffness for the shape retaining cloth (1st row) such that three corners lift off the table mid-fold (2nd row, 2nd column), and high mass/low stiffness result for the heavy, highly deformable polka dot fabric (3rd row) such that all ungrasped corners rest on the table mid-fold (4th row, 2nd column). Right: Compared to data-driven baselines, \textsc{DiffCloud} achieves lower or comparable loss in 13/15 trajectories across categories. We note that all methods struggle to robustly estimate parameters for 2/15 paper towel (shape retaining) manipulation trajectories, which appear as outliers. This is due to difficulties perceiving very thin sheets in motion, which affects all methods.}
    \label{fig:fold}
    \vspace{-15px}
\end{figure}

We find that \textsc{DiffCloud} is able to estimate mass and stiffness cloth parameters that visually explain the observed real point cloud behavior in the lift (Figure \ref{fig:lift}) and fold (Figure \ref{fig:fold}) scenarios. 
We note that \textsc{DiffCloud} achieves a lower loss than \textsc{MeteorNet}, \textsc{PointNet++}, and \textsc{MLP} across the 3 cloth categories in the lift scenario (Figure \ref{fig:lift}), according to the evaluation metric from Section \ref{sec:evaluation}. We also visualize the final parameters found by \textsc{DiffCloud} and observe that the parameters align with the qualitative descriptions of the 3 categories of cloth types considered (Figure \ref{fig:method_timings_params}). For instance, \textsc{DiffCloud} discovers high mass, low stiffness parameters for the polka dot cloth and low mass, high stiffness parameters for the red black cloth, which belong to the highly deformable and shape retaining categories, respectively. Similarly, for the fold scenario, we find that \textsc{DiffCloud} is able to find parameters that account for variations in deformation during execution, such as collapsing inwards mid-fold or maintaining shape (Figure \ref{fig:fold}).
The severity of self-occlusion is more prominent in the fold scenario, making robust parameter estimation more difficult. Still, \textsc{DiffCloud} achieves a loss on par with the baselines, which are (1) trained from thousands of examples, (2) use data augmentations to be invariant to varying degrees of self-occlusion, and (3) require significantly more compute time. \textsc{DiffCloud} takes on average 10 minutes per trajectory in the lift and fold scenarios to optimize the simulation parameters (Figure \ref{fig:method_timings_params}). While the baseline regressors all have inference times on the order of milliseconds, each method incurs an up-front cost of more than two hours of dataset generation. Furthermore, the training procedure requires an additional $40$ minutes to multiple hours per scenario. With significantly less computational footprint, \textsc{DiffCloud} achieves parameter estimation results on real data that are comparable and in some cases better than baselines.

% ----------------------------------
\subsection{Further Simulation Experiments}

\subsubsection{Simulation Scenarios}
\begin{figure}
    \vspace{10px}
    \centering
    \includegraphics[width=0.48\textwidth]{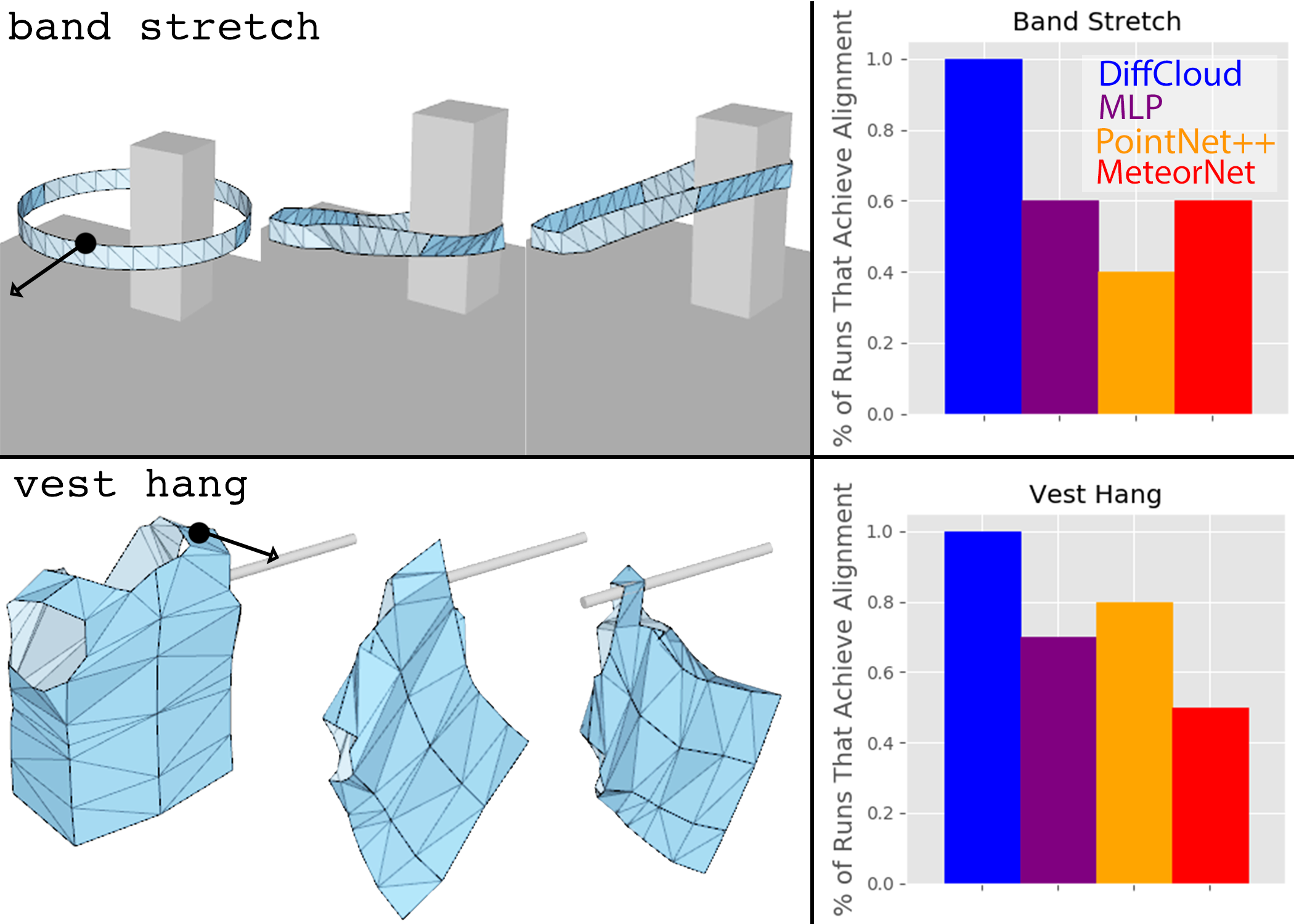}
    \vspace{-22px}
    \caption{We evaluate all competing methods using the metric provided in Section \ref{sec:evaluation} on parameter estimation in two contact-rich, simulated scenarios: \texttt{band stretch} and \texttt{vest hang}. Across 10 runs per scenario, \textsc{DiffCloud} achieves alignment between simulated and target point clouds in all runs, while baseline regressors achieve mixed success. }
    \label{fig:sim_runs}
    \vspace{-20px}
\end{figure}

We further evaluate the performance of \textsc{DiffCloud} on two additional simulated scenarios: hanging one shoulder of a vest onto a pole (\texttt{vest hang}), and stretching an elastic band against a pole (\texttt{band stretch}), shown in Figure~\ref{fig:sim_runs}. The mass and stiffness of the vest mesh determine the outline of the vest when hung on the pole. The mass and stiffness of the elastic band dictate the extent to which the band travels up or down along the pole when pulled taut. We aim to analyze how the contact-rich aspects of these scenarios influence the performance of various methods. To separate these effects from point cloud quality considerations, we use fully observable noise-free point cloud observations. Aside from skipping augmentations (random jittering, dropout of cloth faces) during dataset generation, we use an identical training procedure for \textsc{MLP}, \textsc{PointNet++}, and \textsc{MeteorNet} as in Section \ref{subsec:real_exps}. 

\label{sec:sim_exps}

\subsubsection{\textsc{DiffCloud} Specifications}

\textsc{DiffCloud} performs optimization using the Adam optimizer (with learning rate $0.3$ for \texttt{vest hang}, 0.4 for \texttt{band stretch}). We take the unidirectional Chamfer loss on the last frame for \texttt{band stretch}. For \texttt{vest hang}, most of the deformation happens in the second half of the trajectory, so we pick an intermediate frame in this part of the trajectory (instead of the final frame). We compute the loss between the simulated cloth point cloud only (no poles) and the entire target scene (including poles). This focuses optimization on the deformables. Since synthetic point clouds are noiseless, it is possible to choose a termination criteria for \textsc{DiffCloud} based on when the computed loss falls below a pre-defined threshold. In practice, we find that a loss threshold of $0.0005$ corresponds to a well-aligned match. Thus, for each target trajectory, we run \textsc{DiffCloud} until the loss falls below this threshold or until the number of optimization iterations exceeds $50$, as in the \texttt{lift} and \texttt{fold} scenarios. 

\subsubsection{Results}

For evaluation purposes, we generate a held-out test set of 10 simulated episodes for each scenario with mass and stiffness parameters sampled uniformly at random in the range $[0.1, 10]$. For each episode, we render the corresponding point cloud sequences according to the procedure from Section \ref{sec:grad_prop} without additional augmentations. We evaluate the performance of all methods on estimating the parameters of this set of target point cloud sequences. Across all target point cloud sequences in both scenarios, we find that \textsc{DiffCloud} is able to converge to a set of parameters that yield a Chamfer distance below the threshold. The baselines are only able to achieve sub-threshold alignment in 50-80\% of runs (Figure \ref{fig:sim_runs}). Due to the threshold-based stopping criteria for \textsc{DiffCloud} in the simulated experiments, running optimization for a given target point cloud sequence takes $5$ minutes averaged across the \texttt{vest hang} and \texttt{band stretch} scenarios, compared to a combined dataset generation, training, and inference time on the order of hours for the baselines.

%% file: 5_conclusion.tex
\section{Conclusion and Future Work}

\begin{figure}[t]
    \vspace{10px}
    \centering
    \includegraphics[width=0.48\textwidth]{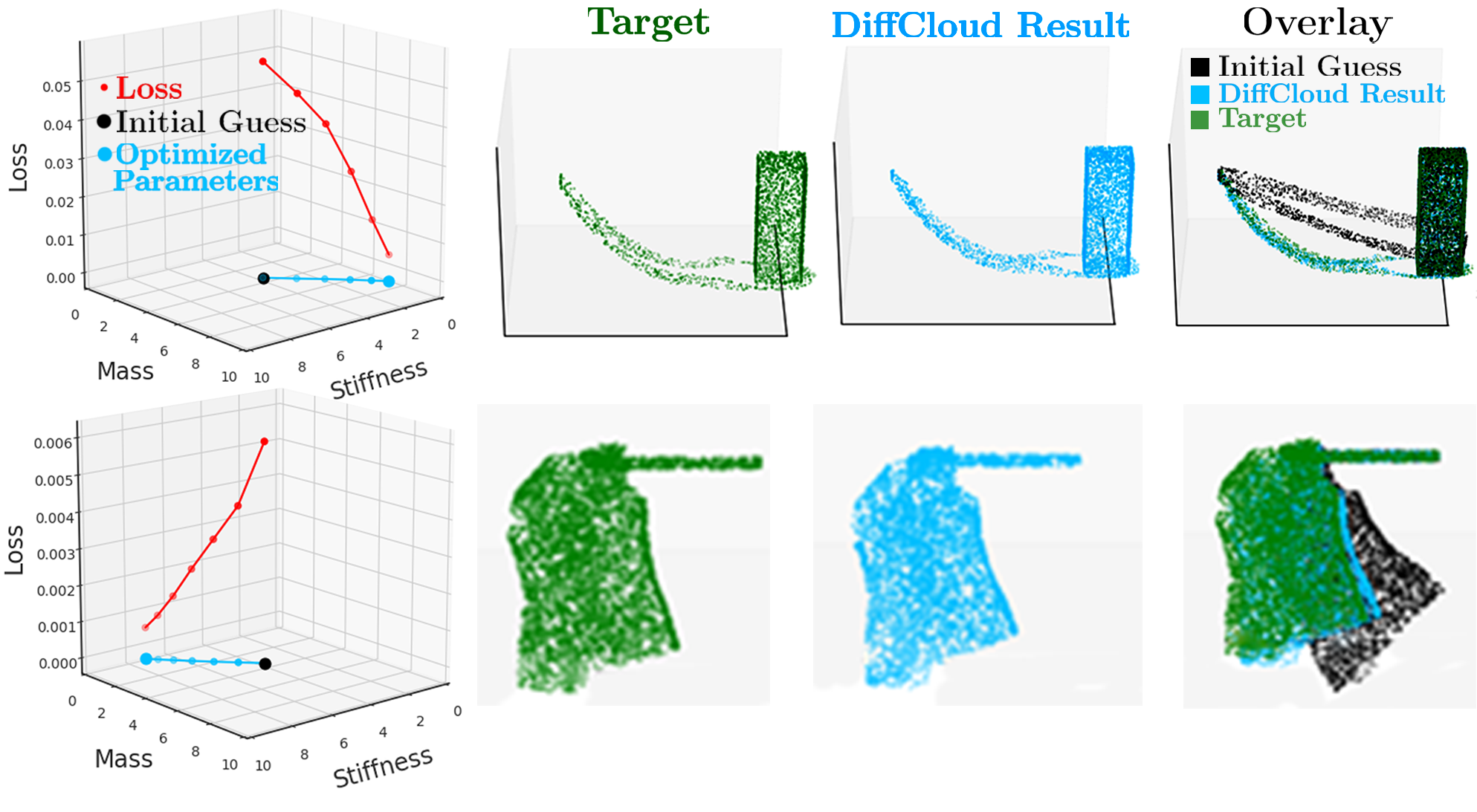}
    \vspace{-25px}
    \caption{\textsc{DiffCloud} optimizes for the mass and stiffness parameters of simulated deformables to match a highly deformable target elastic band (top row) and a shape retaining target vest (bottom row). Starting from an initial guess for the parameters (black), \textsc{DiffCloud} takes gradient steps to update the parameters such that the loss taken between point cloud observations with optimized parameters (blue) and target point clouds (green) is minimized. The optimization terminates once the loss falls below a threshold of 0.0005, denoting close visual alignment.}
    \label{fig:band_vest_loss_params}
    \vspace{-15px}
\end{figure}

In this work, we proposed \textsc{DiffCloud}: an approach to combine differentiable point cloud sampling with differentiable simulation for solving the real-to-sim problems with highly deformable objects. For comparison, we employed recently developed neural network architectures for processing point clouds to learn inverse models that infer simulation parameters without treating the simulator as differentiable. Our experiments showed that \textsc{DiffCloud} reduced the time needed for obtaining real-to-sim alignment by more than an order of magnitude. This result opens the way to more agile experimentation with real-to-sim for highly deformable objects, while still treating the problem as a joint perception-inference task, i.e. not requiring dedicated methods to find correspondences between the depth camera observations (point clouds) and simulated meshes.

Possible avenues for future work include relaxing the need for robot masking by including the robot model into simulation; improving point cloud rendering with realistic occlusion and camera noise models; and jointly optimizing robot actions along with physical simulation parameters. 
Furthermore, it would be interesting to apply \textsc{DiffCloud} to contact-rich multi-stage tasks that involve multiple rigid and deformable objects.
Another interesting direction is to use \textsc{DiffCloud} to infer the starting state of the simulation, including learning the morphology of object meshes. This would allow us to eliminate the need for manual specification of object meshes and initial simulation states, opening the way for creating new realistic simulation scenes `on the fly.'

%
%\it{{\color{gray}A list of our TODOs, some of which could mentioned as future work:
%\begin{itemize}
%    \item Extending to more contact-rich real scenarios, extending to deformables beyond fabrics
%    \item Evaluating found parameters on new trajectories
%    \item Incorporating a point cloud rendering step that uses knowledge of camera extrinsics to render realistic, partial, view-aware point clouds 
%    \item More intelligently using multi-frame / temporal information
%    \item Optimizing for a more diverse range of parameters
%    \item Aligning initial states between sim and real 
%    \item Relaxing the current need for robot masking
%    \item Consider learning an alternative metric for real/sim point cloud alignment or an embedding space for point cloud alignment; could be more informative than Chamfer and less sensitive to outliers
%    \item Using Real2Sim to inform planning
%\end{itemize}
%}
%}